\documentclass{article}

\usepackage{PRIMEarxiv}

\usepackage{amsmath}
\usepackage{algorithm2e}
\usepackage{array, algorithmic, booktabs}

\usepackage[utf8]{inputenc} 
\usepackage[T1]{fontenc}    
\usepackage{hyperref}       
\hypersetup{
    colorlinks=true,
    linkcolor=black,
    citecolor=black,
    urlcolor=black
}
\usepackage{url}            
\usepackage{booktabs}       
\usepackage{amsfonts}       
\usepackage{nicefrac}       
\usepackage{microtype}      
\usepackage{fancyhdr}       
\usepackage{graphicx}       
\graphicspath{{media/}}     

\title{Multiple Additive Neural Networks for Structured and Unstructured Data
}
\date{April 20, 2026}
\author{
\href{https://orcid.org/0000-0001-6450-074X}
    {\hspace{1mm}Jannis Mohr}
    , \href{https://orcid.org/0000-0002-5908-5649}
    {\hspace{1mm}Jörg Frochte}
    \\
  Interdisciplinary Institute for Applied Artificial Intelligence and Data Science Ruhr \\
  Bochum University of Applied Sciences \\
  42579 Heiligenhaus, Germany\\
  \texttt{\{jannis.mohr, joerg.frochte\}@hs-bochum.de} \\
}

\begin{document}
\maketitle

\begin{abstract}
This paper extends and explains the Multiple Additive Neural Networks (MANN) methodology, an enhancement to the traditional Gradient Boosting framework, utilizing nearly shallow neural networks instead of decision trees as base learners. This innovative approach leverages neural network architectures, notably Convolutional Neural Networks (CNNs) and Capsule Neural Networks, to extend its application to both structured data and unstructured data such as images and audio. For structured data the advantages of capsule neural networks as feature extractors are used and combined with MANN as a classifier. MANN's unique architecture promotes continuous learning and integrates advanced heuristics to combat overfitting, ensuring robustness and reducing sensitivity to hyperparameter settings like learning rate and iterations. Our empirical studies reveal that MANN surpasses traditional methods such as Extreme Gradient Boosting (XGB) in accuracy across well-known datasets. This research demonstrates MANN's superior precision and generalizability, making it a versatile tool for diverse data types and complex learning environments.
\end{abstract}


\section{Introduction}
\label{intro}
Boosting combines multiple weak learners to construct a highly accurate model, where each learner only needs to be reasonably precise \cite{Shapire1990}.

Gradient Boosting, a variant of boosting, has shown significant success with structured datasets \cite{Friedman1999,Viola2001}. Typically, this method employs decision trees as the base learners \cite{Chen2016,Dorogush2018}. It operates by using the gradient of one learner to fit the subsequent one, making Gradient Boosting effective at producing highly accurate predictions and less susceptible to overfitting compared to other boosting methods. However, it is still prone to overfitting, especially when several hundred learners are combined \cite{Bikmukhametov2019}.
The MANN algorithm enhances the Gradient Boosting approach by incorporating nearly shallow neural networks as the base learners. This adaptation employs various heuristics and strategies to prevent overfitting, allowing MANN to achieve superior accuracy compared to traditional Gradient Boosting implementations that rely on decision trees. Our development prioritizes ease of use and support for continuous learning, focusing on neural networks with minimal hidden layers and neurons. The MANN algorithm includes mechanisms to automatically halt training if no substantial improvement in accuracy is observed.

This paper aims to advance the discussion on overfitting and adaptivity by demonstrating the effectiveness of using Gradient Boosting with neural networks, achieving superior prediction accuracy over traditional boosting algorithms. Our enhancement focuses on continuous learning to differentiate our approach from standard Gradient Boosting implementations. Continuous learning is crucial for real-world applications, as new data emerges continuously while a model is deployed \cite{Kaeding2017}. MANN is particularly adept at adapting to such scenarios by enabling ongoing training to refine the accuracy of the model on newly acquired data. Our methodology supports two modalities of continuous learning: modifying neural networks within an existing model and using the existing model to compute residuals for fitting a new model, thus creating an enhanced combined framework. Special emphasis is placed on addressing the prevalent issue of overfitting in machine learning. We propose a heuristic that effectively mitigates overfitting risks when applying Gradient Boosting with neural networks, making MANN not only less prone to overfitting but also easier to manage due to reduced hyperparameter tuning compared to other well-known boosting approaches.

Our specific contributions in this paper are summarised below:
\begin{enumerate}
    \item We use the Gradient Boosting algorithm to develop a novel method that incrementally builds deep neural networks out of several neural networks.
    \item We explain heuristics against overfitting and an architecture-based approach for continuous learning.
    \item We show an in-depth analysis of this method with an extensive parameter search and visualise the development of the residuum for an exemplary analytical function.
    \item We adapt this technique to convolutional neural networks with the novel approach of capsule neural networks to train on unstructured data.
    \item We show results on several popular benchmark datasets and compare the results to acclaimed techniques.
\end{enumerate}

In the following section, we provide some related work followed by a detailed description of the algorithm. Section 4 presents the outcomes achieved using our algorithm on the well-known bike-sharing dataset and various classification and regression tasks, along with comparisons against XGBoost, which uses trees as base learners. The paper concludes with a discussion of future research directions and potential applications of MANN.
\section{Related Work} 
\label{sec:RelWork}
In this section an in-depth view on related work is given regarding the main topics of this paper (Boosting, Continuous Learning, Capsule Networks).

\subsection{Adaptations of Boosting Algorithms}
Previous studies have explored the integration of neural networks with boosting techniques. \cite{Schwenk} demonstrated that neural networks could be effectively paired with the Adaptive Boosting (AdaBoost) method, sometimes even outperforming AdaBoost with decision trees. Our research builds on these foundations using the more modern and typically more effective Gradient Boosting framework. The results underscore the robust predictive quality of boosting when combined with neural networks. Moreover, we've incorporated advanced heuristics and designed the algorithm to simplify usage and minimize the need for extensive hyperparameter tuning.

\cite{Martinez-Munoz2019} suggests a methodology for sequentially training the neurons of a neural network with a single hidden layer using a boosting strategy.
\cite{Shalev-Shwartz2014} introduced the SelfieBoost algorithm, which employs Stochastic Gradient Descent as a weak learner within a boosting framework to enhance a single neural network.

Recent research has explored the integration of tree structures with neural networks. Various studies in this area combine tree architectures with neural network elements. Adaptive Neural Trees \cite{Tanno2019} employ a method to progressively develop tree-like structures integrated with neural networks. \cite{Deboleena2018} introduces a concept for assembling hierarchical configurations of multiple neural networks arranged in a tree-like fashion that can expand to accommodate new data.
MANN is distinct from these methods as it leverages a commonly used algorithm in decision tree applications but does not incorporate neural networks into tree-like frameworks.

\subsection{Continuous Learning}
Learning serves as the cornerstone for intelligent systems to effectively adapt to dynamic environments. Over the course of evolution, humans and other organisms have developed remarkable adaptability, continuously acquiring, updating, and utilizing knowledge in response to external changes \cite{Hadsell2020}. Similarly, we expect artificial intelligence systems to exhibit comparable adaptive behaviors. This expectation underpins the study of continual learning, a process where AI systems learn a sequence of contents progressively and manage them as though they were encountered simultaneously. Such contents may include new skills, additional instances of previously learned skills, varying environments, and different contexts, each incorporating its own set of realistic challenges. Given that these contents are provided incrementally throughout an AI system's lifetime continuous learning is frequently synonymous with terms like incremental learning or lifelong learning \cite{Kudithipudi2022}.

A predominant issue in this field is known as catastrophic forgetting \cite{Kirkpatrick.2017}, where an AI's adaptation to a new set of data significantly diminishes its ability to recall previously learned information. This issue represents a critical aspect of the trade-off between learning plasticity and memory stability: an excess of one can adversely affect the other. Ideally, a solution for continual learning should achieve strong generalizability, enabling it to handle distribution variances both within and across different tasks.

Recent advancements in continual learning have been classified into five distinct categories: regularization-based approaches; replay-based approaches that attempt to reconstruct and reuse old data distributions; optimization-based approaches that involve direct manipulations of the optimization processes; representation-based approaches that focus on developing robust and well-distributed representations; and architecture-based approaches that involve designing adaptive parameters tailored to specific tasks \cite{Mohr2021}.

\subsection{Capsule Networks}
Initial studies on a mathematical formulation of the human visual cortex led to groundbreaking work by \cite{GeoffreyE.Hinton.1981}, who observed that certain processes related to human vision have parallels with the concept of inverse computer graphics. In this approach, an image is analyzed with the objective of identifying and locating objects by decomposing complex visual information into simpler geometric figures such as squares and triangles. This concept was further explored in \cite{GeoffreyE.Hinton.1981b}, deepening our understanding of how such methods could potentially mirror some aspects of human cognitive processing in vision.

Building on the idea of inverse computer graphics, subsequent research highlighted that contemporary Convolutional Neural Networks (CNNs) do not operate on this principle and, therefore, differ significantly from human-like vision mechanisms \cite{Tielenman.2014}. In an innovative study, \cite{GeoffreyE.Hinton.2011} presented an initial model for developing neural networks that incorporate elements of inverse computer graphics. This research laid the groundwork for the development of capsule networks, which featured dynamic routing among neuron clusters—termed capsules—and vector sharing across layers, as introduced by \cite{Sabour.2017}. \cite{Hinton.2018} further advanced this model by integrating capsule networks with the Expectation-Maximization algorithm and utilizing matrices for more sophisticated data processing.

The evolution and enhancement of capsule networks have been the subject of extensive research. \cite{Rajasegaran.2019} successfully developed a deeper capsule network that demonstrated improved accuracy on benchmark datasets such as Fashion-MNIST and CIFAR-10. The study by \cite{Xi.2017} specifically analyzed the optimal parameters and network architectures to minimize test errors on CIFAR-10. Further applications and evaluations by \cite{Renkens.2018} revealed that capsule networks could outperform traditional models in specialized applications such as understanding spoken language in command-and-control scenarios. Capsule networks have also been adapted to process 3D input data for action detection in video sequences \cite{Duarte.2018}. Several other studies have confirmed that capsule networks provide superior performance in image classification tasks across various domains compared to baseline CNN models \cite{Afshar.2018,Mobiny.2018,Iesmantas.2018,Kumar.2018,gagana2018activation,abeysinghe2021capsule}. These investigations collectively demonstrate that capsule networks are not only highly effective but also widely applicable in diverse fields.

Moreover, \cite{Zhang.2019} proposes integrating interpretable convolutional filters into CNNs. These filters, which are designed to encode specific parts of an object, could help clarify the decision-making processes of CNNs, somewhat analogous to the information encapsulation found in capsules, albeit requiring explicit integration and targeted training on particular object segments. Additionally, \cite{Simonyan.2013,Mundhenk.2019} suggest using saliency maps (or heatmaps) to identify the focal areas of an image that influence the network's decisions. A novel framework utilizing capsule networks for generating image descriptions offers a more intuitive approach to explain network decisions. However, \cite{Alqaraawi.2020} indicates that while saliency maps provide some insights into the features used by CNNs, they do not necessarily enable users to predict the network's behavior with new images, highlighting a gap that capsule networks could potentially fill.

\section{Gradient Boosting for Neural Networks} \label{sec:AGBNN}
The precision of a predictive model can frequently be enhanced by employing the collective outputs of multiple models, particularly when these individual models are both accurate and distinct. This typically involves using a base learner applied repeatedly across different subsets of the training data.

Gradient Boosting is a methodical approach to assembling predictors in a sequential fashion to develop an additive model, using the gradient of a loss function to guide the training of subsequent predictors in each iteration. In this framework, each predictor contributes equally to the final ensemble, their effects modulated solely by a consistent learning rate, as articulated in the seminal work by \cite{Friedman1999a}.

We consider a system characterized by output variables $y$ and input variables $x$, represented by dataset $\mathcal{D} = {y_i, x_i}$. The objective is to derive a function $F^*(x)$ that maps inputs $x$ to outputs $y$ while minimizing the value of a chosen differentiable loss function $L(y_j,F(x))$, with the index $j$ representing the iteration number $I_j$.

The process commences with an initial approximation $F_0(x)$, serving as the predictive baseline for each $x$. Subsequently, the computation of (pseudo-)residuals is defined as:
\begin{equation}
r_{i,j} = -[\frac{\partial L(y_i, F(x_i))}{\partial F(x_i)}]{F(x)=F{j-1}(x)}
\end{equation}
The term "pseudo-residuals" varies with the loss function, with the residual sum of squares leading to genuine residuals in regression contexts \cite{Friedman1999a}. The base learner is then trained on these $(r_i, x_i)$ pairs, and the outputs $\gamma_{j}$ from this training are calculated. The function update in each iteration is as follows:
\begin{equation}
\gamma_{j}=\operatorname{argmin}{\gamma} \sum{x_i \in R_{i,j}} L(y_i, F_{j-1}(x_i)+\gamma)
\end{equation}
\begin{equation}
F_j(x)=F_{j-1}(x)+\nu \sum_{j=1}^{J} \gamma_j
\end{equation}
This sequence—calculating residuals, fitting the base learner, updating predictions—continues through a predetermined number of iterations $J$, culminating in a fully trained model $\mathcal{M}$, ready for practical deployment.

\begin{algorithm}[tb]
\caption{Multiple Additive Neural Networks}
\label{alg:AGBNN}
\begin{algorithmic}[1]
\STATE {\bfseries Input:} data $x_i, y_i$ ; Loss function $L(y_i, F(x))$
\STATE Initialise $F_0(x) = \operatorname{argmin}\gamma \sum{i=1}^n L(y_i, \gamma )$.
\REPEAT
\STATE $r_{i,j} = -[\frac{\partial L(y_i, F(x_i))}{\partial F(x_i)}]{F(x)=F{j-1}(x)}$
\STATE Fit a neural network to $r_{i,j}$
\STATE Early stopping.
\FOR{$j=1,..., J_j$}
\STATE $\gamma_{j}= \operatorname{argmin}\gamma \sum{x_i in R_{i,j}} L(y_i, F_{j-1}(x_i)+\gamma)$
\ENDFOR
\STATE $F_j(x)=F_{j-1}(x)+\nu \sum_{j=1}^{J_j} \gamma_j $
\STATE Heuristic to prevent overfitting.
\STATE $j = j+1$
\UNTIL{$j = J$}
\end{algorithmic}
\end{algorithm}

The MANN framework adopts neural networks, suitable for both regression and classification tasks exemplified further through innovations like Capsule Networks for image classification. The architecture and training epochs of these neural networks are precisely calibrated to optimize performance and mitigate overfitting, harnessing early stopping mechanisms to fine-tune training duration based on network performance improvements.

\subsection{Integrating a Heuristic to Prevent Overfitting} \label{overfitting}

\begin{algorithm}
\caption{Heuristic to prevent overfitting}
\label{alg:overfitting}
\begin{algorithmic}[1]
   \STATE Let $\mathcal{T}$ be the training set and $\mathcal{V}$ the validation set.
   \STATE Every time training of a neural network $NN_i$ is finished:
   \IF{$E_{va}(NN_i) \geq E_{va}(NN_{i-1}) or \leq E_t$} 
   \STATE break
   \ENDIF
\end{algorithmic}
\end{algorithm}

Preventing Overfitting is a central and important part of every algorithm that fits neural networks on data.

\begin{figure}
\begin{center}
\includegraphics[scale=0.7]{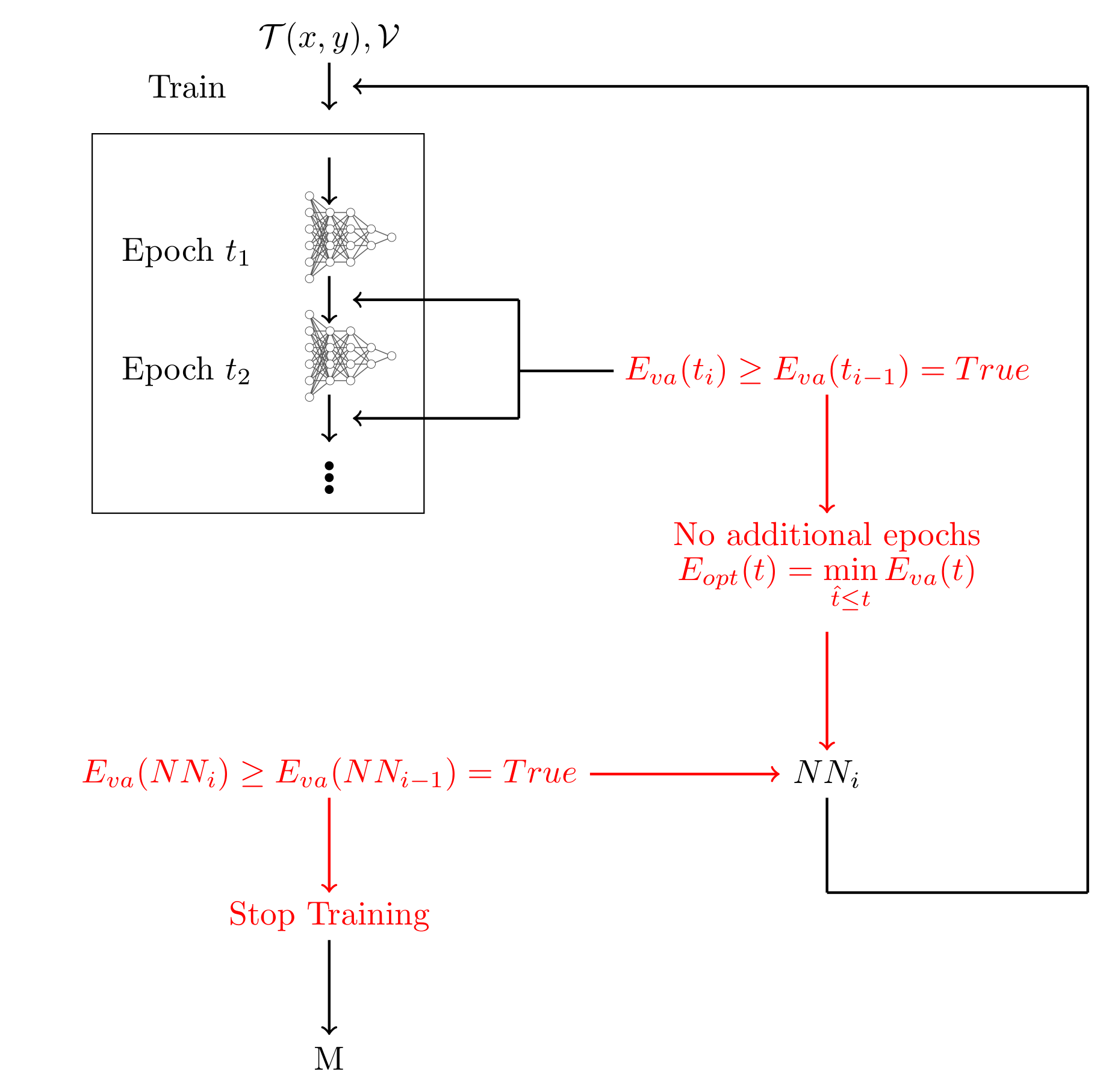}
\caption{Diagrammatic overview of the proposed heuristics against overfitting. The heuristics are active on every iteration of MANN. Shown here is the process of training the neural networks in MANN.}
\label{overfitting_sketch}
\end{center}
\end{figure}

Our heuristic is triggered following the operation at line 10 in Algorithm \ref{alg:AGBNN}, applying its framework at each iteration within the algorithm. A designated segment of the training dataset $\mathcal{T}$ is methodically isolated to establish the validation dataset $\mathcal{V}$, which remains unchanged for the duration of the algorithm and is used to evaluate performance in each iteration. Empirical evaluations across different datasets have shown that allocating around five percent of the training data for validation purposes typically results in optimal performance. The operational dynamics of our heuristic are illustrated in detail in Algorithm \ref{alg:overfitting}.

In each iteration, referred to as iteration $I$, an interim model $\mathcal{M}_1$ is formulated. This model integrates the initial value $F_0$ with the sequentially trained neural networks $NN_j$, reflecting the current state of training and providing an indication of model performance if no further neural networks were added.

The interim model $\mathcal{M}1$ then forecasts outcomes for the $x$ values from the validation dataset. These predictions are rigorously evaluated, and an error metric $E{va}$ is calculated. This error is compared against a pre-set threshold $E_t$. If the error is below this threshold, the model's predictions are considered satisfactory, and the integration of new neural networks is ceased. If, however, the error surpasses the threshold, training continues. This decision-making process is repeated over $n$ iterations, typically three, to assess whether the prediction accuracy remains stable or declines. If accuracy does not improve or worsen across these iterations, the incorporation of additional neural networks is stopped, as further improvements are deemed unlikely.

In line with this approach, an early stopping protocol \cite{Raskutti2013} is employed when training individual neural networks (as noted in Algorithm \ref{alg:AGBNN}). This same validation dataset is consistently used for each evaluation. The performance of each neural network is thoroughly assessed after every training epoch. Training ceases for any neural network that does not show improvement over a specified number of epochs, since continuing the training would likely not enhance performance.

This dual-layer strategy serves to prevent overfitting and ensures that training does not extend beyond a point where it can positively influence the model’s performance. Each neural network is closely monitored and trained for an optimal duration, while the overall model is regularly evaluated at each iteration to maintain a streamlined and efficient structure, thereby minimizing its vulnerability to overfitting.

\subsection{Approach to Continuous Learning for Gradient Boosting with Neural Networks} \label{cl}

\begin{algorithm}
\caption{Continuous Learning with MANN}
\label{alg:continuous}
\begin{algorithmic}[1]
   \STATE Let $\mathcal{T}_1$ and $\mathcal{T}_2$ be two datasets with the same number of features and the same target value.
   \STATE Model $M_1$ trained on $\mathcal{T}_1$ as described in Algorithm~\ref{alg:AGBNN}.
   \IF{$E(M_1(\mathcal{T}_1)) - E(M_1(\mathcal{T}_2)) \geq \epsilon$}
    \STATE break
   \ELSE
    \STATE Retrain Model $M_1$ on the new training data $\mathcal{T}_2$.
    \IF{$|E(M_1(\mathcal{T}_1)) - E(M_1(\mathcal{T}_2))| \geq \epsilon$}
     \STATE break
    \ELSE
     \STATE Algorithm~\ref{alg:AGBNN} with $\mathcal{T}_2$ to build Model $M_2$.
     \STATE Combine $M_1$ and $M_2$ to create final Model $M_3$.
    \ENDIF
   \ENDIF
\end{algorithmic}
\end{algorithm}

A model $M_1$ is trained on a dataset $\mathcal{T}_1$. This model reaches a certain quality of its predictions until the training is stopped. A second dataset $\mathcal{T}_2$ exists with new data, but the same set of features and targets as the first dataset.

First, the algorithm checks if new training is required. The already trained model is fed with the new data and the predictions of the model are evaluated by comparing the error $E$ of the prediction of $\mathcal{T}_1$ with the error $E$ of the prediction of $\mathcal{T}_2$. If the results in a specific metric are close to the original dataset, apparently retraining is not necessary. It is very likely that better results will not be achieved. The two datasets seem to be quite similar.
\begin{equation}
    |E(M_1(\mathcal{T}_1)) - E(M_1(\mathcal{T}_2))| \geq  \epsilon \label{eq:clerror}
\end{equation}

The new dataset is again predicted with the model. If the performance is similar to the original one there is no need to continue the training. Equation \eqref{eq:clerror} is again used for the rating of the performance. $E$ can be an arbitrary error metric for example mean-square-error. If there is still a significant difference the model is extended. Algorithm \ref{alg:AGBNN} is used again but with the new dataset. Generally speaking, new neural networks that are fitted to the new dataset are added to the already existing model $M_1$.

\begin{figure}
\begin{center}
\includegraphics[scale=0.63]{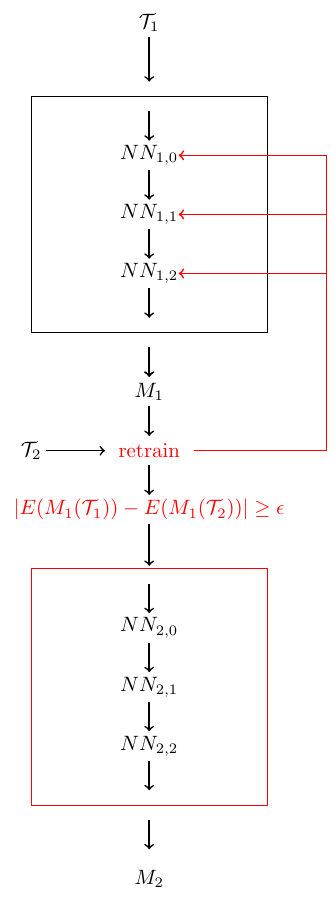}
\caption{Diagrammatic overview of the proposed heuristics against overfitting. The heuristics are active on every iteration of MANN. Shown here is the process of training the neural networks in MANN.}
\label{cl_sketch}
\end{center}
\end{figure}

The algorithm starts off with the initial guess $F_0(x)$. When not training continuously the target values $y_i$ are used for initialisation. Taking advantage of already having a model, the loss of Model $M_1$ on training data $\mathcal{T}_2$ is used as the initial value. Therefore, model $M_1$ is used to start the fitting of new neural networks. From this point on, model $M_2$ is built with neural networks that are fit to the data for a given amount of iterations $I$. 
The final model $M_3$ is then used for predictions. Continuous learning can be repeated as new training data becomes available.
The complete process of continuous learning as pseudo-code is given in Algorithm \ref{alg:continuous} and as a schematic in Figure \ref{cl_sketch}.

\subsection{Integrating Capsule Networks}
\begin{figure}[ht]
\vskip 0.2in
\begin{center}
\centerline{\includegraphics[width=0.8\columnwidth]{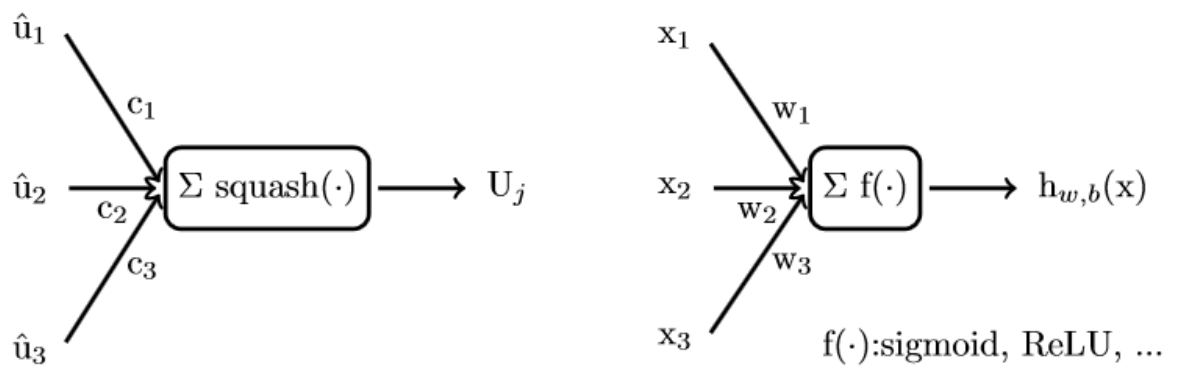}}
\caption{On the left-hand side a capsule can be seen which has in-going vectors and uses the squashing function to get an output vector. On the right-hand side a neuron in the way it is currently used in CNNs can be seen which has an input of scalar values and an activation function (sigmoid, ReLU, ...) to form an output. \cite{Mohr2021}}
\label{capsule_vs_neuron}
\end{center}
\vskip -0.2in
\end{figure}

In the architecture of capsule networks, a single layer is composed of numerous distinct entities known as capsules. Each capsule fundamentally acts as an encapsulating structure that houses a dedicated cluster of neurons. This organizational scheme is visually represented in Figure \ref{capsule_vs_neuron}, which provides a simplified comparison highlighting the structural differences between a traditional neuron and a capsule.

Neurons typically process scalar values, both as input and output. In contrast, capsules encompass a collection of neurons that operate together to produce vector outputs from vector inputs. The adoption of vectors allows capsule networks to encapsulate and preserve rich image details such as location and color attributes within their outputs. The magnitude of these output vectors is particularly significant; it denotes the probability of the presence of a specific feature within the input, thereby maintaining crucial image information without loss. Each n-dimensional capsule is capable of learning and subsequently outputting an n-dimensional vector. To ensure that these vectors effectively represent probabilities, their magnitudes are constrained to values between 0 and 1. To achieve this, a novel non-linear squashing function was developed, which is detailed in Equation \ref{eq:squash}. Here, $v_j$ represents the vector output from capsule $j$, and $s_j$ denotes the total input to that capsule \cite{Sabour.2017}.

The design of a capsule network typically culminates in a final layer comprising as many capsules as there are distinct classes within the dataset. Each capsule’s output vector provides a rich, multi-dimensional representation of the class-specific image attributes and the likelihood of each class being represented in the input image. These dimensions articulate various attributes associated with each class, allowing the network to differentiate and recognize a diverse array of features inherent in complex data inputs. This capability not only enhances the interpretability of the network’s decisions but also significantly improves its ability to generalize from training data to novel, unseen datasets. This structural organization and functional methodology underscore the sophistication and potential of capsule networks in handling intricate pattern recognition tasks in machine learning.

\begin{equation} \label{eq:squash}
    v_{j} = \frac{||s_j||^2}{1+||s_j||^2} \times \frac{s_j}{||s_j||}
\end{equation}

In the scope of inverse graphics a way to connect e. g. a detected eye to a possible face and not to a car is needed. This is done by a novel dynamic routing technique called routing-by-agreement. With routing-by-agreement the output of a capsule is directed to a higher-level capsule where it agrees with other inputs. For example several detected facial features like a mouth and a nose should be routed to a capsule that detects faces. Instead, a detected mouth and tire should not be directed to the same high-level capsule and therefore not agree with each other.
It is very unlikely for agreements to lie close to each other in a high-dimensional space (coincidence filtering). Thus, an accumulation can not be a coincidence and, therefore, strengthens the decision of the network. \cite{Sabour.2017}

The aforementioned Capsule Networks are utilized as base learners for image classification with gradient boosting in the same fashion neural networks are used for structured data in the proposed MANN algorithm with slight adaptations. Capsule Networks are used because they work especially well on small datasets and due to their characteristics might benefit from a boosting approach. Their added decoder network additionally makes it possible to generate images which can be used for continuous learning in a generative kind of way or to make the trained models more robust. Throughout this paper an architecture which is based on the architecture proposed by
Sabour et al. \cite{Sabour.2017} is used.

section{Experiments}
In the ensuing section, we present a series of experiments designed to evaluate the accuracy and operational mechanics of our proposed machine learning method, specifically highlighting the utility of our heuristics in combating overfitting. These experiments utilize an academic implementation of the Multiple Additive Neural Networks (MANN) model. For regression tasks, we employ the loss function $L(y_j,F(x)) = \frac{1}{2} \cdot (y_j - F(x))^2$, which measures the squared differences between the predicted values and the actual values. For classification tasks, the loss function used is $L(y_j,p) = y_i \cdot \log(p) + (1-y_i) \cdot \log(1-p)$, where p represents the predicted probability of the positive class. This function is essential for evaluating the performance of the model in probabilistic terms, providing a robust measure of classification accuracy.

The neural networks implemented within the MANN framework consist of three hidden layers, each comprising eight neurons. This configuration was deliberately chosen to maintain consistency across experiments, thereby minimizing parameter variation and enhancing the comparability of results. By standardizing the network architecture, we aim to isolate the effects of our methodological innovations and ensure that the observed outcomes are attributable to the efficacy of the MANN approach rather than differences in network structure.

To further solidify the reliability of our findings, each experiment was repeated multiple times. This extensive testing is crucial to ascertain that the selected parameters consistently lead to high performance across various runs, thereby mitigating any anomalies or outliers in the results.

For comparative analysis, we employ Extreme Gradient Boosting (XGB), utilizing its widely accessible implementation. XGB serves as a benchmark for assessing the performance enhancements provided by our method, given its reputation and widespread use in tackling similar tasks.

Additionally, in the regression scenarios, we included experiments using a multi-layer perceptron (MLP) with five layers. This setup is representative of typical deep neural network architectures without the application of boosting techniques. The inclusion of MLP allows us to directly compare the impacts of deep architectures against those augmented with our boosting-based approach, providing a clear perspective on the advantages and limitations of employing boosting in complex neural network models.

Adaptive Neural Trees (ANT) is an innovative technique that combines the structural and algorithmic modalities of decision trees with the representational power of neural networks to form a versatile learning framework \cite{Tanno2019}. This approach leverages the hierarchical nature of decision trees and the deep learning capabilities of neural networks to create a model that is both interpretable and capable of handling complex, high-dimensional data.
The core idea behind ANT is to utilize a decision tree-like architecture where each node can perform a specific function that either splits the data based on certain criteria or processes the data using neural network components. This architecture allows the model to adaptively grow and adjust its complexity based on the requirements of the task at hand, making it particularly effective for tasks where the relationship between input features and outputs is non-linear and intricate. ANT is used as an example for novel approaches which use tree-like structures for adding neural networks while training.

This structured approach to experimentation not only tests the robustness and effectiveness of the MANN model but also carefully contrasts it against established methodologies, thus offering comprehensive insights into its practical applications and potential superiority in specific machine learning tasks.

\subsection{Fitting an Analytical Function}

\begin{figure}[ht]
\vskip 0.2in
\begin{center}
\includegraphics[width=\columnwidth]{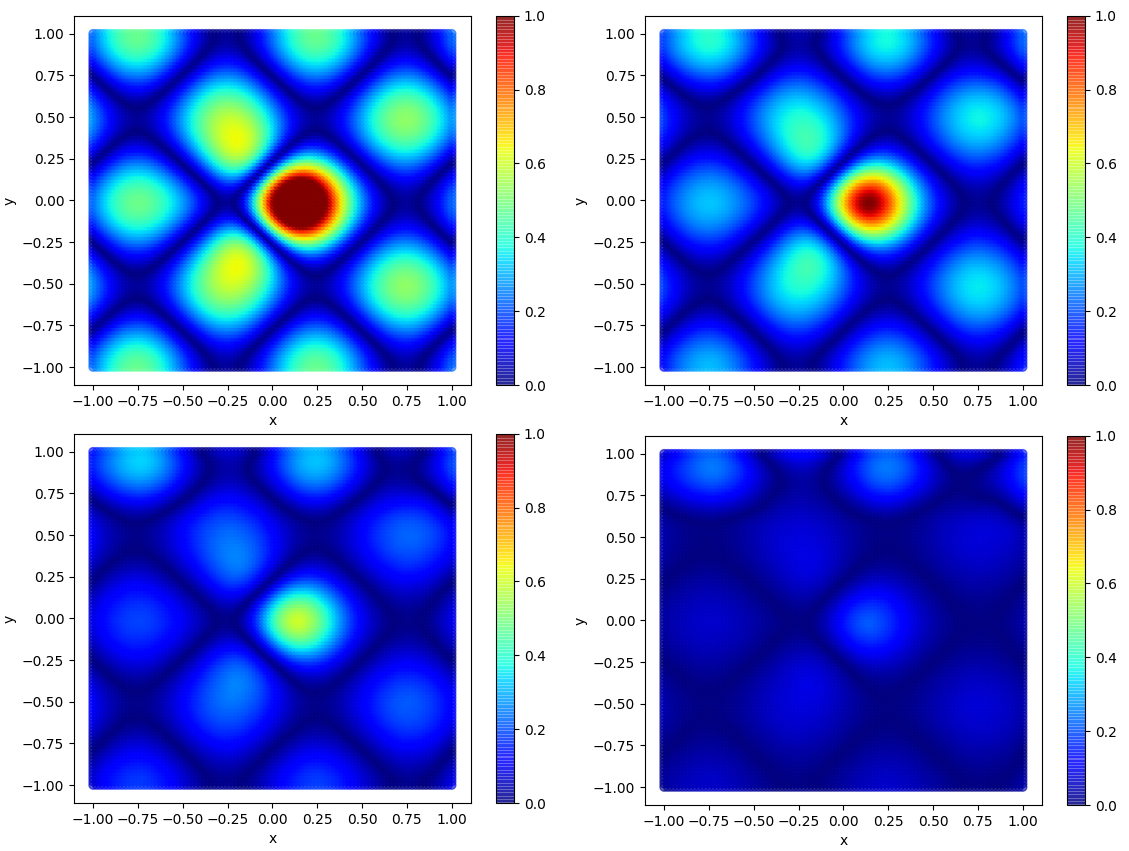}
\caption{Development of the residuum while fitting a MANN model on the analytical function described in \eqref{eq:analytical_function}. The figure shows the residuum before training the first neural network and after 5, 10 and 15 iterations (from top left to bottom right). The scale reaches from blue to red with red being the largest residuum. It can be seen that the residuum gradually decreases with every iteration.}
\label{analytical_fnc_error_plot}
\end{center}
\vskip -0.2in
\end{figure}

To visualise the development of the residuum a model was trained with our algorithm on an analytical function. The function 
\begin{equation}
\begin{split}
f(x,y)=(0.25 + 0.75 \cdot e^{( -10 \cdot (x^2+y^2))}) \\
\cdot (\sin(2 \cdot \pi \cdot x)+\cos(2 \cdot \pi \cdot y))
\label{eq:analytical_function}
\end{split}
\end{equation}
was used.
This function has a peak and is not highly symmetric. The dataset is generated with x and y values in the interval $[-1;1]$ and 5\% gaussian noise is used on the target values.

In this experiment, the sigmoid activation function and the Adam optimiser are used. The learning rate is set to 0.1. 20 iterations of the algorithm were made and therefore the final model consists of 20 artificial neural networks. Due to our heuristics to prevent overfitting every neural network was trained in a different number of epochs and with a maximum number of 100 epochs. The dataset was generated with the given equation and with a grid of $100 \times 100$ samples. The values of the function are the target values and the input variables $x\text{ and }y$ are the features.

Figure \ref{analytical_fnc_error_plot} shows the development of the absolute residuum plotted in a 2D plot that shows the function in a top-down view.  The residuum before the first iteration is the mean of all values (this is only true because the squared loss is used) in the dataset. It is the initial guess $F_0(x)$.

\begin{table}[t]
\caption{Metrics on the analytical function.}
\label{Analytical_metric}
\vskip 0.15in
\begin{center}
\begin{small}
\begin{sc}
\begin{tabular}{lccc}
\toprule
Algorithm & MAE & MSE & RMSE \\
\midrule
MANN & 0.040 & 0.0035 &  0.059\\
XGB & 0.075 & 0.016 & 0.125 \\
ANT & 0.048 & 0.0052 & 0.067 \\
\bottomrule
\end{tabular}
\end{sc}
\end{small}
\end{center}
\vskip -0.1in
\end{table}

From there on, the error of the model gets smaller and smaller. The edges of the deep red area are focused first, while the error on predicting the function is being reduced as a whole.
Using the proposed algorithm with the parameters previously named leads to a mean-squared error of 0.0035 (mean-absolute error = 0.040) on the whole dataset. XGB, with a learnrate of 0.3 and 200 trees with a maximum depth of 3, has a mean-squared error of 0.16 (mean-absolute error = 0.075) on this dataset.

\subsection{Results on Regression Benchmarks}
The following sections contain a study on the performance on several regression benchmarks to show the accuracy of MANN compared to other popular methods.

\subsubsection{Results on the Bike Sharing Dataset}

\begin{figure*}
\begin{center}
\includegraphics[scale=0.5]{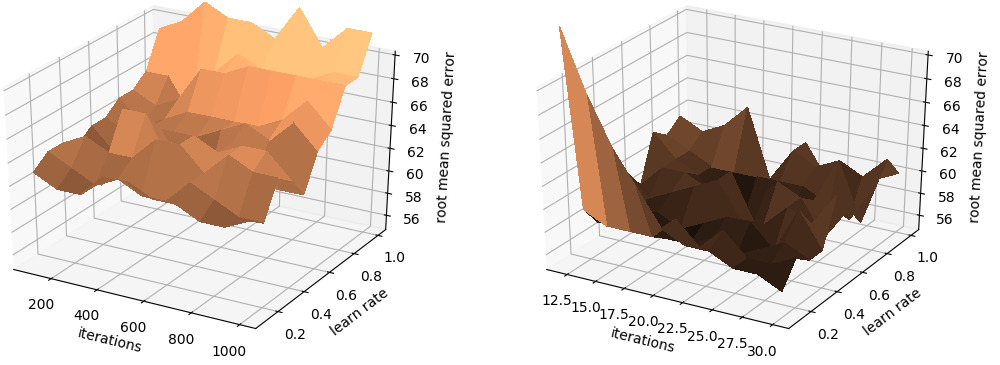}
\caption{RMSE plotted depending on the number of iterations and learn rate training the bike sharing dataset with XGB (left) and MANN (right). This plot represents the parameter grid search that was used to find the parameters for MANN and XGB that lead to the best accuracy. The root mean squared error is used as a metric. Furthermore, this plot shows that MANN is less dependent on its parameters then XGB. \cite{Mohr2023}}
\label{BikeSharing_eval_surface}
\end{center}
\end{figure*}

We utilized the Bike Sharing dataset, a well-regarded dataset released in 2013 that has been widely used for algorithm testing and educational purposes \cite{Fanaee-T2014}. This dataset logs detailed information on each bicycle rental, including duration, start and end dates, start and end stations, bike number, and membership type of the user. Additionally, it is enriched with environmental data such as weather conditions, humidity, temperature, wind speed, and the day of the week, encompassing hourly data over a span of two years, resulting in 17,379 records, each containing 15 features.

No data augmentation or outlier removal was performed. We conducted a parameter grid search to identify the optimal combination of learning rate and number of iterations, capped at 500 epochs. Stochastic gradient descent served as the optimizer, and the training of new networks was discontinued once the mean absolute error on the validation dataset fell below 1. The learning rate was set at 0.3, achieving a root mean squared error (RMSE) of 56 in predicting the test data, which comprised data from the last several days of each month, starting on the 20th, to simulate a realistic use-case scenario.

For comparative purposes, an XGB model was also trained on this dataset with a learning rate of 0.2 and a maximum tree depth of 6, resulting in an RMSE of 62 on the test data. Figure \ref{BikeSharing_eval_surface} depicts the RMSE fluctuations over various iterations and learning rates for both the MANN and XGB models.

\subsubsection{SARCOS Multivariate Regression}

The SARCOS regression dataset is a highly complex regression task originally proposed by \cite{Vijazakumar2000}. It represents the learning of the inverse dynamics of a seven-degrees-of-freedom robot arm. The task is to map from a 21-dimensional input space (7 joint positions, 7 joint velocities, 7 joint accelerations) to the corresponding joint torques. The dataset has 44484 training examples and 4449 test examples. The benchmark data are publicly available.
All best-performing learners on this dataset are tree-based and outperform ANNs by some margin. Additionally, most learners have a penchant for overfitting or underfitting on this rather small dataset. An ensemble of 8 ANTs achieves an MSE as low as 1.226 on the SARCOS dataset \cite{Tanno2019}.
MANN is the first learner that achieves an MSE close to tree-based learners while not overfitting on this problematic dataset. The settings described in the introduction of this section were again used for this experiment. It achieves an MSE of 1.429 which is very close to the 1.444 from XGB but a bit worse than the 1.226 that an ensemble of ANTs achieves. This also is another indication that MANN is quite secure from overfitting even on small datasets which easily make ANNs overfit.
A Multi-Layer Perceptron with 5 hidden layers is used as a representation for learners based on neural networks and what they can achieve on this dataset. As shown by \cite{Tanno2019} neural networks with more layers do not further lower the error but increase it because they overfit on this dataset.

All results can be seen in table \ref{table:regression}.

\subsubsection{Million Song Dataset \& CT Scan Slize Localization Dataset}

The CT Scan Slice Localization Dataset, introduced by \cite{Graf2011}, encompasses medicinal data derived from 53,500 CT images collected from 74 patients (43 male, 31 female). This dataset features 384 distinct attributes extracted from these images. The construction of the feature vector involves the concatenation of two histograms: one representing the positioning of bone structures and another detailing the presence of air inclusions. The dataset's target values span a numeric range from 0 to 180.

The Million Song Dataset (MSD), detailed by \cite{Bertin-Mahieux2011}, offers a comprehensive compilation of audio features for songs recorded between 1922 and 2011. The challenge associated with this dataset involves predicting the year a song was released based solely on its audio features, excluding any metadata. To ensure a rigorous evaluation, the dataset's authors recommend a specific division into training and test sets that prevents the same artists from appearing in both sets, thereby avoiding data leakage. Additionally, the distribution of data across different years within the dataset is notably non-uniform, presenting further challenges for predictive modeling. All results can be seen in table \ref{table:regression}.

\begin{table}[t]
\caption{Regression Datasets (CT Scan and MSD in RMSE, SARCOS in MSE). Extended from \cite{Mohr2023}.}\label{table:regression}
\begin{center}
\begin{tabular}{lcccc}
\toprule
Dataset & MANN & XGB  & ANT & MLP\\
\midrule
CT Scan & 5.34 & 6.67 & - & 8.49\\
MSD & 8.57 & 9.38 & - & 12.73\\
SARCOS & 1.43 & 1.45 &  1.23 & 2.66\\
\bottomrule
\end{tabular}
\end{center}
\end{table}

\subsection{Results on Binary Classification Datasets}
As described in Section \ref{sec:AGBNN} MANN does not only support regression but also classification tasks. Four binary classification datasets were selected to present a variety of complexity and dataset sizes and the accuracy of MANN, XGB, and ANT are compared. Results can be seen in table \ref{table:classification}.\\

\subsubsection{UCI Heart Disease \& Rain in Australia \& Titanic Dataset}

The UCI Heart Disease Dataset, introduced by \cite{Detrano1989}, comprises four subsets sourced from different hospitals, documenting patient data both with and without heart disease. This dataset aims to ascertain the presence and type of heart disease in patients, featuring 14 variables across 303 instances. While some studies \cite{Dangare2012,Zriqat2016} report near-perfect accuracy using extensive data augmentation, the majority \cite{Chen2011,Sabarinathan2014,Sabay2018} achieve around 85 percent accuracy without such techniques. For consistency, no data augmentation was employed in this experiment. A learning rate of 0.6 was set, and 18 neural networks were each trained for up to 400 epochs. With these settings, MANN achieved a 90 percent accuracy rate, surpassing XGB's 85 percent and slightly exceeding the typical outcomes reported in most related literature \cite{Aljanabi2018}, demonstrating MANN's efficacy on small datasets.

The Rain in Australia dataset comprises daily meteorological data from various locations across Australia, with the objective of predicting next-day rainfall. It includes 23 features over 142,193 records, collected from multiple weather stations over a decade, sourced from the Australian Bureau of Meteorology. Forecasting rain is critically important due to its implications for agriculture and as an early indicator for natural disasters \cite{Parmar2017}. The feature $RISK_MM$, indicating the subsequent day's rainfall, was omitted from the training data to ensure target integrity. The dataset contains several categorical features, such as wind direction, which were processed using Label Encoding.
Optimal parameters were identified via grid search, setting the learning rate parameter grid as $M_{\nu} = {\nu \in N| 2 \leq \nu \leq 20 \land \nu = 2 \times k }$ for both algorithms. XGB was run for 600 iterations, showing consistent accuracy with a peak at 87 percent using a learning rate of 0.4. MANN, however, performed slightly better, achieving a top accuracy of 89 percent with a learning rate of 0.5 over 18 iterations and averaging an accuracy of 84.87 percent. The ANT model matched MANN's performance at 89 percent on this predictive task.

The Titanic dataset remains one of the most engaged datasets within the machine learning community, largely due to the historical significance and human element associated with the events surrounding the Titanic tragedy \cite{Ekinci2018}. The primary dataset includes demographic and travel-related details of 891 passengers in the training set and 418 in the test set. The main challenge posed to researchers is to predict whether a passenger survived the sinking of the ship based on variables including their name, sex, and cabin, among other features.

The dataset presents certain complexities, such as missing values; for some passengers, critical information such as ticket price or number of children is absent.
In an experiment, a model comprising 10 neural networks trained over 500 epochs was developed, with a learning rate of 0.15 that yielded the most favorable outcomes. Published studies \cite{Singh2017,Rajesh2019} report accuracy scores ranging from 80\% to 86\% using various methodologies. With the aforementioned parameters, our model, MANN, achieved an accuracy of 84\%, which, while not surpassing the leading classifiers, remains competitive. Notably, the XGB model outperforms MANN on this dataset, attaining an accuracy score of 0.86, aligning it with the highest-rated classifiers. The relative underperformance of MANN can be attributed to its typical proficiency with larger datasets, as indicated by other experiments. This suggests that MANN may not be ideally suited for smaller datasets, a conclusion that highlights the need for adaptive strategies depending on the size and complexity of the dataset.

\begin{table}[t]
\caption{Binary Classification Datasets Accuracy. Extended from \cite{Mohr2023}.}\label{table:classification}
\begin{center}
\begin{tabular}{lccc}
\toprule
Dataset & MANN & XGB & ANT\\
\midrule
UCI Heart Disease & 0.90 & 0.85 & 0.88\\
Titanic & 0.84 & 0.86 & 0.85\\
Rain in Australia & 0.89 &  0.87 & 0.89\\
Higgs Boson & 0.85 & 0.83 & 0.82\\
\bottomrule
\end{tabular}
\end{center}
\end{table}

\subsubsection{Higgs Boson Dataset}
The Higgs Boson dataset presents a sophisticated classification challenge, aimed at distinguishing between events that are likely to result in the production of Higgs Bosons and those that are not. Originating from Monte Carlo simulations, this dataset meticulously replicates the stochastic behavior observed in particle physics experiments. It incorporates a range of features, including kinematic properties and various derived metrics that are standard in physics research. With over 1 million entries, the dataset exemplifies the extensive scale and depth of data typically encountered in high-energy physics investigations \cite{Baldi_2014}.

This dataset serves as a testing ground for the MANN model, particularly in handling large and imbalanced datasets, where events not resulting in Higgs Boson production vastly outnumber those that do. The outcomes of this assessment are documented in table \ref{table:classification}, which outlines the performance metrics attained by MANN. These results highlight the model’s strengths and limitations within the context of large-scale and unbalanced data environments. Such analyses are vital for discerning how various learning models fare under different data conditions and contribute to broader discussions concerning the appropriateness of models across various scientific fields.

\subsection{Continuous Learning Benchmark}
We propose a novel split of this dataset to use it as a benchmark for continuous learning on structured regression tasks. This dataset was strategically divided into two distinct segments corresponding to the years 2011 and 2012, each containing similar features. This division is justified by the clear structural delineation present within the data, suggesting a natural split. On average, the dataset shows an increase in bike rentals in 2012 compared to 2011. 

We first trained a model, designated as $M_1$, using the MANN algorithm with previously specified parameters, on the dataset from 2011. The model achieved a root-mean-squared error (RMSE) of 57 when predicting for 2011. However, when this model was applied to predict the 2012 dataset, the RMSE significantly increased to 128, highlighting the challenges of model generalization over time.

To address this, we employed our continuous learning strategy as outlined in section \ref{cl}, where the neural networks of model $M_1$ were further trained using the 2012 data. This adjustment reduced the RMSE to 106, demonstrating the initial impact of continuous learning. Subsequently, we enhanced the model by training additional neural networks specifically on the 2012 data and integrating them into the existing model framework. This second level of continuous learning further improved the RMSE to 79, reflecting the robustness of our approach in adapting to new data while retaining previously learned patterns.

For comparative analysis, we also employed Extreme Gradient Boosting (XGB) trained solely on the 2011 data, which achieved an RMSE of 58 for that year and 130 when applied to the 2012 dataset. Additionally, an artificial neural network (ANN) with five hidden layers and a neuron configuration of 20, 15, 10, 5, and 1, trained over 700 epochs, was tested. This network achieved an RMSE of 69 for 2011 and 155 for 2012. The ANN was then retrained on the 2012 dataset, which improved its RMSE to 92 for that year, indicating some benefits of retraining, but not to the extent observed with our continuous learning approach.

Furthermore, we explored the capabilities of Learn++.MT, an algorithm based on AdaBoost with decision trees and specifically designed for incremental learning as proposed by \cite{Muhlbaier2004}. The results from all these experiments, including a comprehensive performance comparison, are detailed in table \ref{table:bs_cl}. This analysis underscores the superiority of our MANN model with its continuous learning framework in adapting to new data, significantly reducing prediction errors compared to traditional models and even advanced machine learning techniques designed for incremental learning.

\begin{table}
\caption{Bike Sharing Comparison RMSE \cite{Mohr2023}}
\label{table:bs_cl}
\begin{center}
\begin{small}
\begin{sc}
\begin{tabular}{lccc}
\toprule
Algorithm & 2011 & 2012 & both \\
\midrule
MANN frozen & 57 & 128 & 56\\
MANN cl 1st level & 57 & 106 & 56\\
MANN cl all levels & 57 & 79 & 56\\
XGB frozen & 58 & 130 & 62 \\
Learn++.MT & 60 & 87 & 63\\
ANN frozen & 69 & 155 & 67\\
ANN cl & 69 & 92 &  67\\
\bottomrule
\end{tabular}
\end{sc}
\end{small}
\end{center}
\end{table}

\subsection{Image Classification Benchmark}

\begin{table}[t]
\caption{Image Classification Benchmark Accuracy}
\label{table:icb}
\begin{center}
\begin{small}
\begin{sc}
\begin{tabular}{lcc}
\toprule
Algorithm & MNIST & CIFAR-10 \\
\midrule
plain capsule network \cite{Sabour.2017} & 96.6 & 89.4\\
MANN & 99.1 & 91.8\\
CNN & 98.8 & 88.2\\
\bottomrule
\end{tabular}
\end{sc}
\end{small}
\end{center}
\end{table}

The capsule network is used without reconstruction and 3 routing iterations. In addition to that, the architecture and parameters are used as described by \cite{Sabour.2017}. For comparison, the results for a plain capsule network with the same parameters but without any boosting Furthermore, a simple CNN with Dropout, MaxPooling and 6 Convolutional layers, with 128 neurons is used. All results can be seen in table \ref{table:icb}. Overall, it can be observed that MANN with capsule networks achieves a slight advantage over the datasets used in terms of accuracy over simple CNNs and plain capsule networks. This is promising for future work and can be extended to networks with more layers to achieve better results.

\subsubsection{MNIST}
The MNIST dataset is a cornerstone in the field of machine learning, often serving as a benchmark for evaluating the performance of various image processing systems \cite{deng2012}. Originally derived from the NIST (National Institute of Standards and Technology) datasets, MNIST was modified by LeCun et al. to be more suitable for machine learning applications. It consists of a collection of 70,000 handwritten digits, divided into a training set of 60,000 images and a test set of 10,000 images. Each image is a grayscale bitmap of size 28x28 pixels, representing a single digit from 0 to 9.
The simplicity and size of MNIST make it an ideal dataset for training and testing algorithms on tasks such as classification, recognition, and machine learning introductory exercises. Over the years, MNIST has played a pivotal role in the development and comparison of models in the neural networks domain, particularly with the advent and growth of deep learning techniques.
Furthermore, the use of MNIST in numerous studies has provided the machine learning community with a standardized benchmark for evaluating and comparing the performance of different learning algorithms. 

\subsubsection{CIFAR-10}
The CIFAR-10 dataset is a fundamental resource in the field of machine learning, particularly in the study and development of image recognition algorithms. Comprising a collection of 60,000 32x32 color images, this dataset is divided into a training set of 50,000 images and a test set of 10,000 images. The images are uniformly distributed across ten different classes, which include airplanes, automobiles, birds, cats, deer, dogs, frogs, horses, ships, and trucks. Each class is represented by 6,000 images, making the dataset balanced in terms of class distribution.

Originally published by Krizhevsky \cite{Krizhevsky2009}, CIFAR-10 is derived from the '80 million tiny images' dataset and serves as a stripped-down version for quick testing of algorithms. This dataset challenges researchers due to its relatively low resolution and the high degree of visual similarity between classes, such as those between cats and dogs, or trucks and automobiles.

\section{Conclusion}
In this paper, we have presented a novel integration of the Gradient Boosting algorithm with neural networks, particularly focusing on the MANN algorithm for enhancing predictive accuracy and addressing the challenges of overfitting and continuous learning in machine learning models. Our study extends the capability of Gradient Boosting by employing neural networks as base learners, moving beyond the traditional use of decision trees. This adaptation not only leverages the structural advantages of neural networks, but also incorporates various heuristics to effectively mitigate overfitting, thereby enhancing model performance and stability across diverse datasets.

Through extensive experiments and evaluations, we have demonstrated that the MANN algorithm not only achieves higher accuracy in comparison to traditional Gradient Boosting models but also excels in environments requiring adaptability to new data. The introduction of continuous learning mechanisms further allows MANN to dynamically update and improve its predictions in response to new information, a critical feature for real-world applications where data continuously evolves.

We have shown that the integration of shallow neural networks within the Gradient Boosting framework can lead to substantial improvements in prediction accuracy. We have introduced and validated a heuristic that successfully prevents overfitting, which is particularly valuable when dealing with complex models like those involving neural networks. In addition, we used Capsule Networks to enhance MANN with the potential to adapt to tasks like image classification.

Future research directions will explore deeper and more complex neural network architectures within the MANN framework, evaluate the scalability of our approach to larger datasets, and further refine our heuristics to enhance model interpretability and management. We also plan to extend our continuous learning framework to include more sophisticated data streaming scenarios, enabling our models to be more responsive to rapid changes in data characteristics.

Overall, the findings from our research provide a solid foundation for advancing the field of machine learning by improving the robustness, accuracy, and flexibility of models designed for both structured and unstructured data environments. This will undoubtedly contribute to the development of more intelligent and adaptable machine learning systems capable of handling the complexities and dynamics of real-world data.

\section*{Acknowledgments}
This work was funded by the Federal Ministry of Education and Research under 16-DHB-4021.

\bibliographystyle{unsrt}  
\bibliography{references}

@article{abeysinghe2021capsule,
  title={Capsule Networks for Character Recognition in Low Resource Languages},
  author={Abeysinghe, C and Perera, I and Meedeniya, DA},
  journal={Machine Vision Inspection Systems, Volume 2: Machine Learning-Based Approaches},
  pages={23--46},
  year={2021},
  publisher={Wiley Online Library}
}

@inproceedings{gagana2018activation,
  title={Activation function optimizations for capsule networks},
  author={Gagana, B and Athri, HA Ujjwal and Natarajan, S},
  booktitle={2018 International Conference on Advances in Computing, Communications and Informatics (ICACCI)},
  pages={1172--1178},
  year={2018},
  organization={IEEE}
}

@article{GeoffreyE.Hinton.1981,
 author = {{Geoffrey E. Hinton}},
 year = {1981},
 title = {{A Parallel Computation that Assigns Canonical Object-Based Frames of Reference}},
 journal = {Seventh International Joint Conference on Artificial Intelligence}
}

@article{GeoffreyE.Hinton.1981b,
 author = {{Geoffrey E. Hinton}},
 year = {1981},
 title = {{Shape Representation in Parallel Systems}},
 journal = {Seventh International Joint Conference on Artificial Intelligence}
}

@article{Hadsell2020,
  title={Embracing Change: Continual Learning in Deep Neural Networks},
  author={Raia Hadsell and Dushyant Rao and Andrei A. Rusu and Razvan Pascanu},
  journal={Trends in Cognitive Sciences},
  year={2020},
  volume={24},
  pages={1028-1040},
}

@article{GeoffreyE.Hinton.2011,
 author = {{Geoffrey E. Hinton} and {A. Krizhevsky} and {S. Wang}},
 year = {2011},
 title = {{Transforming Auto-Encoders}},
 journal = {International Conference on Artificial Neural Networks}
}

@misc{Krizhevsky2009,
  title={Learning Multiple Layers of Features from Tiny Images},
  author={Alex Krizhevsky},
  year={2009}
}

@article{Hinton.2018,
 author = {Hinton, Geoffrey E. and Sabour, Sara and Frosst, Nicholas},
 year = {2018},
 title = {{Matrix capsules with EM routing}},
 journal = {{International Conference on Learning Representations}},
}

@misc{Sabour.2017,
 author = {Sabour, Sara and Frosst, Nicholas and Hinton, Geoffrey E.},
 year = {2017},
 title = {{Dynamic Routing Between Capsules}},
 journal = {Conference on Neural Information Processing Systems}
}

@phdthesis{Tielenman.2014,
 author = {Tielenman, Tijmen},
 year = {2014},
 title = {{Optimizing Neural Networks That Generate Images}},
 address = {Toronto},
 publisher = {Department of Computer Science},
 school = {{University of Toronto}},
 type = {{Dissertation}}
}

@misc{Iesmantas.2018,
 author = {Iesmantas, Tomas and Alzbutas, Robertas},
 year = {2018},
 title = {{Convolutional capsule network for classification of breast cancer histology images}}
}

@misc{Mobiny.2018,
 author = {Mobiny, Aryan and {van Nguyen}, Hien},
 year = {2018},
 title = {{Fast CapsNet for Lung Cancer Screening}}
}

@article{Afshar.2018,
 author = {Afshar, Parnian and Mohammadi, Arash and Plataniotis, Konstantinos N.},
 year = {2018},
 title = {{Brain Tumor Type Classification via Capsule Networks}},
 journal = {25th IEEE International Conference on Image Processing ICIP}
}

@misc{Zhang.2019,
 author = {Zhang, Quanshi and Wang, Xin and Wu, Ying Nian and Zhou, Huilin and Zhu, Song-Chun},
 year = {2019},
 title = {{Interpretable CNNs for Object Classification}}
}

@misc{Mundhenk.2019,
 author = {Mundhenk, T. Nathan and Chen, Barry Y. and Friedland, Gerald},
 year = {2019},
 title = {{Efficient Saliency Maps for Explainable AI}}
}

@article{Simonyan.2013,
 author = {Simonyan, Karen and Vedaldi, Andrea and Zisserman, Andrew},
 year = {2013},
 title = {{Deep Inside Convolutional Networks: Visualising Image Classification Models and Saliency Maps}},
 journal = {International Conference on Learning Representations ICLR}
}

@article{Alqaraawi.2020,
 author = {Alqaraawi, Ahmed and Schuessler, Martin and Wei{\ss}, Philipp and Costanza, Enrico and Berthouze, Nadia},
 year = {2020},
 title = {{Evaluating Saliency Map Explanations for Convolutional Neural Networks: A User Study}},
 journal = {International Conference on Intelligent User Interfaces}
}

@article{Kumar.2018,
 author = {Kumar, Amara Dinesh},
 title = {{Novel Deep Learning Model for Traffic Sign Detection Using Capsule Networks}},
 year = {2018},
 journal = {{International Journal of Pure and Applied Mathematics Volume 118 No. 20}}
}

@article{Duarte.2018,
 author = {Duarte, Kevin and Rawat, Yogesh S. and Shah, Mubarak},
 year = {2018},
 title = {{VideoCapsuleNet: A Simplified Network for Action Detection}},
 journal = {Advances in Neural Information Processing Systems}
}

@article{Renkens.2018,
 author = {Renkens, Vincent and {van hamme}, Hugo},
 year = {2018},
 title = {{Capsule Networks for Low Resource Spoken Language Understanding}},
 journal = {Proc. Interspeech 2018}
}

@article{Xi.2017,
 author = {Xi, Edgar and Bing, Selina and Jin, Yang},
 year = {2017},
 title = {{Capsule Network Performance on Complex Data}},
 journal = {International Joint Conference on Neural Networks (IJCNN)}
}

@article{Rajasegaran.2019,
 author = {Rajasegaran, Jathushan and Jayasundara, Vinoj and Jayasekara, Sandaru and Jayasekara, Hirunima and Seneviratne, Suranga and Rodrigo, Ranga},
 year = {2019},
 title = {{DeepCaps: Going Deeper with Capsule Networks}},
 journal = {IEEE/CVF Conference on Computer Vision and Pattern Recognition CVPR}
}

@misc{Kudithipudi2022,
author = "Dhireesha Kudithipudi and Mario Aguilar-Simon and Jonathan Babb and Maxim Bazhenov and Douglas Blackiston and Josh Bongard and Andrew P Brna and Suraj Chakravarthi Raja and Nick Cheney and Jeff Clune and Anurag Daram and Stefano Fusi and Peter Helfer and Leslie Kay and Nicholas Ketz and Zsolt Kira and Soheil Kolouri and Jeffrey L Krichmar and Sam Kriegman and Michael Levin and Sandeep Madireddy and Santosh Manicka and Ali Marjaninejad and Bruce McNaughton and Risto Miikkulainen and Zaneta Navratilova and Tej Pandit and Alice Parker and Praveen K Pilly and Sebastian Risi and Terrence J Sejnowski and Andrea Soltoggio and Nicholas Soures and Andreas S Tolias and Darío Urbina-Meléndez and Francisco J Valero-Cuevas and Gido M van de Ven and Joshua T Vogelstein and Felix Wang and Ron Weiss and Angel Yanguas-Gil and Xinyun Zou and Hava Siegelmann",
title = "{Biological underpinnings for lifelong learning machines}",
year = "2022",
month = "3",
}

@article {Kirkpatrick.2017,
	author = {Kirkpatrick, James and Pascanu, Razvan and Rabinowitz, Neil and Veness, Joel and Desjardins, Guillaume and Rusu, Andrei A. and Milan, Kieran and Quan, John and Ramalho, Tiago and Grabska-Barwinska, Agnieszka and Hassabis, Demis and Clopath, Claudia and Kumaran, Dharshan and Hadsell, Raia},
	title = {Overcoming catastrophic forgetting in neural networks},
	volume = {114},
	number = {13},
	pages = {3521--3526},
	year = {2017},
	doi = {10.1073/pnas.1611835114},
	publisher = {National Academy of Sciences},
	issn = {0027-8424},
	eprint = {https://www.pnas.org/content/114/13/3521.full.pdf},
	journal = {Proceedings of the National Academy of Sciences}
}

@misc{Martinez-Munoz2019,
  author   = {Gonzalo Martinez-Munoz},
  title    = {Sequential Training of Neural Networks with Gradient Boosting},
  year     = {2019},
}

@InProceedings{Graf2011,
author="Graf, Franz
and Kriegel, Hans-Peter
and Schubert, Matthias
and P{\"o}lsterl, Sebastian
and Cavallaro, Alexander",
editor="Fichtinger, Gabor
and Martel, Anne
and Peters, Terry",
title="2D Image Registration in CT Images Using Radial Image Descriptors",
booktitle="Medical Image Computing and Computer-Assisted Intervention -- MICCAI 2011",
year="2011",
publisher="Springer Berlin Heidelberg",
address="Berlin, Heidelberg",
pages="607--614",
isbn="978-3-642-23629-7"
}

@misc{Muhlbaier2004,
author = {Muhlbaier, Michael and Topalis, Apostolos and Polikar, Robi},
year = {2004},
pages = {52-61},
title = {Learn++.MT: A New Approach to Incremental Learning},
volume = {3077}
}

@article{Deboleena2018,
author = {Deboleena, Roy and Priyadarshini, Panda and Kaushik, Roy},
year = {2018},
title = {Tree-CNN: A Hierarchical Deep Convolutional Neural Network for Incremental Learning},
journal = {Neural Networks}
}

@article{Vijazakumar2000,
author = {Vijayakumar, Sethu and Schaal, Stefan},
year = {2000},
title = {Locally Weighted Projection Regression: An O(n) Algorithm for Incremental Real Time Learning in High Dimensional Space},
journal = {Proceedings of the Seventeenth International Conference on Machine Learning (ICML 2000)}
}

@Article{Tanno2019,
  author  = {Ryutaro Tanno and Kai Arulkumaran and Daniel Alexander and Antonio Criminisi and Aditya Nori},
  title   = {Adaptive Neural Trees},
  journal = {7th International Conference on Advanced Technologies},
  year    = {2018},
}

@Article{Ekinci2018,
  author  = {Ekin Ekinci and Sevinc Ilhan Omurca and Neytullah Acun},
  title   = {A Comparative Study on Machine Learning Techniques Using Titanic Dataset},
  journal = {36th International Conference on Machine Learning},
  year    = {2019},
}

@Article{Rajesh2019,
  author   = {Rajesh M},
  title    = {Predictions of Survivors in the Titanic Cruise},
  journal  = {International Journal of Recent Technology and Engineering},
  year     = {2019},
  doi      = {10.35940/ijrte.c4408.098319},
}

@Article{Singh2017,
  author   = {Aakriti Singh and Shipra Saraswat and Neetu Faujdar},
  title    = {Analyzing Titanic Disaster using Machine Learning Algorithms},
  journal  = {International Congerence on Computing, Communication and Automation},
  year     = {2017},
  doi      = {10.1109/ccaa.2017.8229835},
}

@InProceedings{Parmar2017,
  author   = {Parmar, Aakash and Mistree, Kinjal and Sompura, Mithila},
  title    = {Machine Learning Techniques For Rainfall Prediction: A Review},
  booktitle = {International Conference on Innovations in information Embedded and Communication Systems},
  year     = {2017},
}

@Article{Bikmukhametov2019,
  author   = {Timur Bikmukhametov and Johannes Jaeschke},
  title    = {Oil Production Monitoring using Gradient Boosting Machine Learning Algorithm},
  journal  = {12th IFAC Symposium on Dynamics and Control of Process Systems},
  year     = {2019},
}

@Article{Chen2011,
  author   = {Austin S. Chen and Shao-Wei Huang and Paul S. Hong and CC Cheng and Edward J. Lin},
  title    = {HDPS: Heart disease predictions system},
  journal  = {Computing in Cardiology},
  year     = {2011},
}

@Article{Sabarinathan2014,
  author   = {V. Sabarinathan and V. Sugumaran},
  title    = {Diagnosis of Heart Disease using Decision Trees},
  journal  = {International Journal of research in Computer Applications and Information Technology},
  year     = {2014},
  pages    = {74-79},
}

@Article{Zriqat2016,
  author   = {Esraa Zriqat and Ahmad Altamimi and Mohammad Azzeh},
  title    = {A Comparative Study for Predicting Heart Diseases Using Data Mining Classification Methods},
  journal  = {International Journal of Computer Science and Information Security},
  year     = {2016},
  volume   = {12},
}

@Article{Dangare2012,
  author  = {Chaitrali S. Dangare and Sulabha S. Apte},
  title   = {A Data Mining Approach for Predictions of Heart Disease Using Neural Networks},
  journal = {International Journal of Computer Engineering and Technology},
  year    = {2012},
  pages   = {30-40},
}

@Article{Sabay2018,
  author       = {Alfeo Sabay and Laurie Harris and Vivek Bejugama and Karen Jaceldo-Siegl},
  title        = {Overcoming Small Data Limitations in Heart Disease Prediction by Using Surrogate Data},
  journal      = {SMU Data Science Review},
  year         = {2018},
  volume       = {1},
  number       = {3},
}

@Article{Aljanabi2018,
  author       = {Maryam Aljanabi and Mahmoud H. Qutqut and Mohammad Hijjawi},
  title        = {Machine Learning Classification Techniques for Heart Disease Prediction: A Review},
  journal      = {International Journal of Enigneering and Technology},
  year         = {2018},
}

@Article{Detrano1989,
  author   = {Robert Detrano and Andras Janosi and Walter Steinbrunn and Matthias Pfisterer and Johann-Jakob Schmid and Sarbjit Sandhu and Kern H. Guppy, Stella Lee and Victor Froehlicher},
  title    = {International application of a new probability algorithm for the diagnosis of coronary artery diseas},
  journal  = {The American Journal of Cardiology},
  year = {1989},
}

@InProceedings{Kaeding2017,
  author    = {Kaeding, Christoph and Rodner, Erik and Freytag, Alexander and Denzler, Joachim},
  title     = {Fine-Tuning Deep Neural Networks in Continuous Learning Scenarios},
  booktitle = {Computer Vision - ACCV 2016 Workshops},
  year      = {2017},
  pages     = {588-605},
}

@misc{Raskutti2013,
  author      = {Garvesh Raskutti and Martin J. Wainwright and Bin Yu},
  title       = {Early stopping and non-parametric regression: An optimal data-dependent stopping rule},
  year        = {2013},
}

@Article{Fanaee-T2014,
  author       = {Fanaee-T, H. and Gama, J.},
  title        = {Event labeling combining ensemble detectors and background knowledge},
  journal      = {Progress in Artificial Intelligence},
  year         = {2014},
  pages        = {113-127},
}

@Article{Chen2016,
  author  = {Tianqi Chen and Carlos Guestrin},
  title   = {XGBoost: A Scalable Tree Boosting System},
  journal = {KDD 16},
  year    = {2016},
}

@Article{Friedman1999,
  author       = {Jerome H. Friedman},
  title        = {Stochastic Gradient Boosting},
  journal      = {Computational Statistics \& Data Analysis},
  year         = {1999},
  volume       = {38},
}

@Article{Friedman1999a,
  author  = {Jerome H. Friedman},
  title   = {Greedy Function Approximation: A Gradient Boosting Machine},
  journal = {The Annals of Statistics},
  year    = {1999},
}

@Article{Dorogush2018,
  author      = {Anna Veronika Dorogush and Vasily Ershov and Andrey Gulin},
  title       = {CatBoost: gradient boosting with categorical features support},
  journal     = {Conference on Neural Information Processing Systems (NeurIPS 2018)},
  year        = {2018},
}

@Article{Schwenk,
  author   = {Holger Schwenk and Yoshua Bengio},
  title    = {Boosting Neural Networks},
  journal  = {Neural Computation},
  year     = {2000},
}

@INPROCEEDINGS{Bertin-Mahieux2011,
  author = {Thierry Bertin-Mahieux and Daniel P.W. Ellis and Brian Whitman and Paul Lamere},
  title = {The Million Song Dataset},
  booktitle = {{Proceedings of the 12th International Conference on Music Information
	Retrieval ({ISMIR} 2011)}},
  year = {2011}
}

@article{Baldi_2014,
	year = 2014, 
	publisher = {Springer Science and Business Media {LLC}},
	author = {P. Baldi and P. Sadowski and D. Whiteson},  
	title = {Searching for exotic particles in high-energy physics with deep learning},  
	journal = {Nature Communications}
}

@misc{Shalev-Shwartz2014,
  author      = {Shai Shalev-Shwartz},
  title       = {SelfieBoost: A Boosting Algorithm for Deep Learning},
  year        = {2014},
}

@article{deng2012,
  title={The mnist database of handwritten digit images for machine learning research},
  author={Deng, Li},
  journal={IEEE Signal Processing Magazine},
  volume={29},
  number={6},
  pages={141--142},
  year={2012},
  publisher={IEEE}
}

@conference{Mohr2021,
author={Janis Mohr. and Basil Tousside. and Marco Schmidt. and JÃ¶rg Frochte.},
title={Explainability and Continuous Learning with Capsule Networks},
booktitle={Proceedings of the 13th International Joint Conference on Knowledge Discovery, Knowledge Engineering and Knowledge Management - KDIR,},
year={2021},
pages={264-273},
publisher={SciTePress},
organization={INSTICC},
HIDEdoi={10.5220/0010681300003064},
isbn={978-989-758-533-3},
issn={2184-3228},
}

@conference{Mohr2023,
author={Janis Mohr and Basile Tousside and Marco Schmidt and Jörg Frochte},
title={Multiple Additive Neural Networks: A Novel Approach to Continuous Learning in Regression and Classification},
booktitle={Proceedings of the 15th International Joint Conference on Computational Intelligence - Volume 1: NCTA},
year={2023},
pages={540-547},
publisher={SciTePress},
organization={INSTICC},
HIDEdoi={10.5220/0012234000003595},
isbn={978-989-758-674-3},
}

@Article{Shapire1990,
  author   = {Robert E. Shapire},
  title    = {The strength of weak learnability},
  journal  = {Machine Learning},
  year     = {1990},
  pages    = {197-227},
}

@Article{Viola2001,
  author  = {Paul Viola and Michael Jones},
  title   = {Rapid Object Detection using a Boosted Cascade of Simple Features},
  journal = {Conference on Computer Vision and Pattern Recognition},
  year    = {2001},
}

\end{document}